\documentclass[10pt,twocolumn,letterpaper]{article}

\usepackage[pagenumbers]{cvpr} 

\usepackage{colortbl}

\usepackage{colortbl}

\definecolor{cvprblue}{rgb}{0.21,0.49,0.74}
\usepackage[pagebackref,breaklinks,colorlinks,allcolors=cvprblue]{hyperref}

\def\paperID{*****} 
\def\confName{CVPR}
\def\confYear{2026}

\title{Synthetic Designed Experiments for Diagnosing Vision Model Failures}

\author{Krisanu Sarkar\\
Indian Institute of Technology Bombay\\
Mumbai, India
}

\begin{document}
\maketitle
\begin{abstract}
Current synthetic data pipelines for computer vision generate images without diagnosing what the downstream model actually needs. This open-loop paradigm treats synthetic data as cheap real data---randomly sampling the generator's output space and hoping for coverage of the model's failure modes. We argue that this fundamentally misuses synthetic data's unique property: controllable, independent variation of scene factors. Drawing on the statistical theory of \emph{Design of Experiments} (DoE)---a principled framework for efficiently probing complex systems that, to our knowledge, has not been applied to synthetic data generation for vision---we propose \textbf{Synthetic Designed Experiments for Representational Sufficiency (SDRS)}. SDRS treats the downstream model as a black-box system under investigation and the synthetic generator as an experimental apparatus. Using fractional factorial designs, SDRS efficiently audits a model's factor-sensitivity profile via ANOVA decomposition, classifying failures into two actionable types: \emph{Type~I gaps} (coverage failures on underrepresented factor levels) and \emph{Type~II gaps} (reliance on spurious nuisance dependencies). The audit then prescribes targeted synthetic data to address each gap type. We validate SDRS on three experiments: (1)~a controlled diagnostic on dSprites with planted biases, where the audit correctly identifies both gap types and targeted data improves accuracy from 49.9\% to 79.0\%; (2)~a dense segmentation task on procedural scenes, where the audit detects background-complexity shortcuts and targeted data improves mIoU from 0.948 to 0.998; and (3)~an entanglement detection experiment showing that the ANOVA audit identifies cross-factor contamination in imperfect generators ($\Delta F = 4.7$ for the leaked factor). We further show that per-factor invariance penalties can transfer sensitivity between factors, identifying an open problem for representation-level correction. Code will be released.
\end{abstract}
\section{Introduction}
\label{sec:intro}

The synthetic data pipeline for computer vision has converged on a standard recipe: generate large volumes of images using a controllable source (a 3D renderer~\cite{tremblay2018training,greff2022kubric} or a conditioned diffusion model~\cite{rombach2022high,zhang2023adding}), optionally filter for quality, and add them to the training set alongside real data. This paradigm has demonstrated consistent gains across object detection~\cite{chen2023geodiffusion}, semantic segmentation~\cite{xu2023open}, depth estimation~\cite{ke2024repurposing}, and human pose recovery~\cite{black2023bedlam}.

Yet a fundamental question remains underexplored: \emph{which} synthetic images should be generated? Current pipelines answer this question with heuristics---text prompts chosen by the researcher, domain randomization that uniformly samples all factor combinations, or difficulty scores derived from a pretrained classifier~\cite{komejisatori2024difficulty}. These approaches are \textbf{open-loop}: they do not use the downstream model's specific failure modes to guide generation. When the downstream model already handles 95\% of the visual distribution competently, the vast majority of randomly generated synthetic images provide zero learning signal. This is a waste of computation and, more importantly, a missed opportunity.

We observe that controllable synthetic generators---whether 3D engines or layout-conditioned diffusion models---share a critical structural property: they expose \emph{independent, named parameters} for scene factors such as lighting, viewpoint, material, and occlusion. This makes them functionally equivalent to the experimental apparatuses studied in the statistical theory of \emph{Design of Experiments} (DoE)~\cite{fisher1935design,box2005statistics}, a framework developed precisely for the problem of efficiently understanding how a complex system responds to multiple controllable variables.

The connection is direct. In classical DoE, an experimenter varies input factors according to a structured plan (e.g., a fractional factorial design) and analyzes the system's output via ANOVA to decompose response variance into contributions from individual factors and their interactions. This structured approach is exponentially more sample-efficient than random sampling---a result established by Fisher in 1935~\cite{fisher1935design} and foundational to experimental science. To our knowledge, this connection has never been exploited in the synthetic data literature.

We propose \textbf{Synthetic Designed Experiments for Representational Sufficiency (SDRS)}, a framework that applies DoE principles to diagnose and address vision model failures. SDRS operates in four phases:

\begin{enumerate}
\item \textbf{Designed Experiment.} Generate a small, structured set of synthetic images by varying scene factors according to a fractional factorial design---requiring as few as $2^{k-p}$ images for $k$ factors, rather than the full $\prod_i l_i$ combinations.

\item \textbf{Representational Audit.} Pass these images through the downstream model and perform ANOVA on the per-image task loss, decomposing loss variance into per-factor contributions. This produces a \emph{factor-sensitivity profile} that reveals exactly which factors the model's predictions depend on.

\item \textbf{Gap Diagnosis.} Cross-reference the sensitivity profile with the known task structure. Nuisance factors with significant dependence are \textbf{Type~II gaps} (spurious shortcuts). Factors with strong performance degradation on underrepresented or unseen levels are \textbf{Type~I gaps} (coverage failures). This taxonomy is exhaustive: every factor falls into one of four quadrants.

\item \textbf{Targeted Prescription.} Generate synthetic data that specifically addresses each diagnosed gap: diverse samples along Type~I factors to build missing capability, and matched counterfactual pairs along Type~II factors to enable invariance regularization. Optionally, re-audit after training to verify convergence.
\end{enumerate}

This framework makes three contributions:

\noindent\textbf{(1) A principled diagnostic for synthetic data.} The ANOVA-based audit provides a decomposed, per-factor assessment of what a model has and has not learned. This directly answers the workshop's call for ``benchmark and evaluation methods for synthetic data'' by turning the synthetic generator into a structured evaluation instrument.

\noindent\textbf{(2) A unifying lens on existing approaches.} We show that domain randomization~\cite{tobin2017domain}, counterfactual data augmentation~\cite{mao2021generative}, and active learning~\cite{settles2009active} emerge as special cases of SDRS under specific assumptions in the SDRS framework, and that each corresponds to a suboptimal experimental design.

\noindent\textbf{(3) Empirical validation and an identified open problem.} Across three experiments---a controlled classification task on dSprites, a dense segmentation task on procedural scenes, and an entanglement detection study---we validate the diagnostic and demonstrate that data targeted by the audit dramatically outperforms no-synthetic baselines. We also identify a \emph{sensitivity transfer} phenomenon: per-factor invariance penalties can suppress one shortcut while amplifying others, suggesting that holistic representation constraints are needed for the correction phase.

\section{Related Work}
\label{sec:related}

SDRS sits at the intersection of synthetic-data generation, representation diagnostics, and experimental design.

\paragraph{Domain randomization and sim-to-real.}
Domain randomization (DR)~\cite{tobin2017domain,tremblay2018training} improves robustness by sampling nuisance factors broadly, and has been successful in sim-to-real pipelines~\cite{sadeghi2017cadrl,james2019sim}. However, DR is open-loop: it does not diagnose which factors actually drive downstream errors. SDRS adds a designed audit step that decomposes failure by factor before prescribing corrections.

\paragraph{Counterfactual and invariance-based methods.}
Counterfactual augmentation and invariance objectives~\cite{mao2021generative,arjovsky2019invariant} aim to suppress spurious correlations. SDRS is complementary: it first identifies which nuisance factors are problematic, then targets those factors during correction. This reduces dependence on manual factor selection.

\paragraph{Uncertainty-guided generation and data selection.}
Active learning and uncertainty-guided generation~\cite{settles2009active,zhu2017generative,komejisatori2024difficulty} score sample informativeness at the sample level. SDRS instead provides \emph{factor-level attribution}: it explains why uncertainty arises by decomposing error sensitivity across controllable factors.

\paragraph{Distillation and probing.}
Dataset distillation/condensation methods~\cite{wang2018dataset,zhao2021dataset,cazenavette2023generalizing} optimize compact synthetic sets but are typically expensive and trajectory-dependent. Probing methods~\cite{alain2016understanding,hewitt2019designing,belinkov2022probing} analyze encoded information but are usually descriptive. SDRS uses ANOVA on task loss as a prescriptive diagnostic that directly determines what data to generate next.
\section{Framework}
\label{sec:framework}

We formalize SDRS as an audit-and-prescription loop for downstream model $f_w$ and controllable generator $G$. The generator is parameterized by discrete factors $\mathbf{z}=(z_1,\ldots,z_k)$; factors may be semantic ($\mathbf{z}_S$) or nuisance ($\mathbf{z}_N$).

\subsection{Phase 1: Designed Experiment}
\label{sec:designed_exp}

Instead of random sampling, we use a fractional factorial design. For $k$ two-level factors, a Resolution~IV design~\cite{box2005statistics} uses $2^{k-p}$ runs (versus $2^k$ full factorial). Example: for $k{=}5$, a $2^{5-2}_{\mathrm{IV}}$ plan uses 8 probe points while preserving unconfounded main effects.

Given designed settings $\mathcal{E}=\{\mathbf{z}^{(1)},\ldots,\mathbf{z}^{(M)}\}$, we generate probe data
\begin{equation*}
\mathcal{D}_{\text{probe}}=\{(x^{(j)},y^{(j)},\mathbf{z}^{(j)})\}_{j=1}^{M},\qquad x^{(j)}=G(\mathbf{z}^{(j)}).
\end{equation*}

\subsection{Phase 2: Representational Audit via ANOVA}
\label{sec:audit}

For each probe sample we compute task loss $\ell^{(j)}=\mathcal{L}(f_w(x^{(j)}),y^{(j)})$ (scalarized per image for dense prediction). For each factor $z_j$, we run one-way ANOVA over grouped losses and compute
\begin{equation}
\label{eq:anova}
F_j = \frac{\mathrm{MS}_{\text{between}}(z_j)}{\mathrm{MS}_{\text{within}}(z_j)}.
\end{equation}

To control multiple comparisons, we apply Holm--Bonferroni correction over the $k$ factor tests within each audit pass and report adjusted significance at $\alpha{=}0.05$.

\textbf{Why ANOVA on task loss rather than linear probing.}\quad Linear probes~\cite{alain2016understanding,hewitt2019designing} measure extractability from representations, whereas ANOVA on task loss measures prediction-level dependence, which is the operational quantity for failure diagnosis.

\subsection{Phase 3: Gap Diagnosis}
\label{sec:diagnosis}

We use two explicit tests per factor:
\begin{itemize}[leftmargin=*]
\item \textbf{Coverage test (Type~I candidate):} stratified performance drop on underrepresented or unseen levels (estimated on an audit-validation split).
\item \textbf{Shortcut test (Type~II candidate):} nuisance-factor dependence detected by ANOVA ($z_j\in\mathbf{z}_N$ with Holm-adjusted significance).
\end{itemize}

Classification priority is deterministic: if the coverage test is positive, assign \textbf{Type~I}; else if shortcut test is positive, assign \textbf{Type~II}; else assign \textbf{Correct}. This resolves overlap cases (e.g., a nuisance factor that is both under-covered and sensitive) by prioritizing coverage correction first.

\vspace{0.4em}
\begin{center}
\small
\resizebox{0.95\columnwidth}{!}{%
\begin{tabular}{l|c|c}
\toprule
& \textbf{Shortcut+} & \textbf{Shortcut--} \\
\midrule
\textbf{Coverage+} & \cellcolor{red!15}\textbf{Type~I} & \cellcolor{red!15}\textbf{Type~I} \\
\textbf{Coverage--} & \cellcolor{red!15}\textbf{Type~II} & \cellcolor{green!15}Correct \\
\bottomrule
\end{tabular}%
}
\end{center}
\vspace{0.4em}

\subsection{Phase 4: Targeted Prescription}
\label{sec:prescription}

\paragraph{Type~I correction (coverage restoration).}
For each Type~I factor, generate diversity-focused synthetic data over the deficient levels while sampling remaining factors from the training distribution.

\paragraph{Type~II correction (shortcut suppression).}
For each Type~II factor, generate matched counterfactual pairs $(x_a,x_b)$ with identical semantics and all nuisance factors fixed except $z_j$, then enforce representation invariance.

\begin{equation}
\label{eq:inv_loss}
\mathcal{L}_{\text{inv}}=\frac{1}{|\mathcal{P}|}\sum_{(a,b)\in\mathcal{P}}\frac{1}{d_l}\left\|\Phi_w^{(l)}(x_a)-\Phi_w^{(l)}(x_b)\right\|_2^2,
\end{equation}
where $d_l$ is the number of elements in layer-$l$ features (e.g., $d_l{=}H_lW_lC_l$ for dense maps), making regularization scale explicit across architectures.

\paragraph{Combined objective.}
\begin{equation}
\label{eq:total_loss}
\mathcal{L}_{\text{total}}=\mathcal{L}_{\text{task}}(\mathcal{D}_{\text{real}}\cup\mathcal{D}_{\text{Type~I}})+\lambda\,\mathcal{L}_{\text{inv}}(\mathcal{P}_{\text{Type~II}}),
\end{equation}
with $\lambda{=}0.5$ in all experiments unless stated otherwise.

\paragraph{Verification.}
After each correction round, rerun the audit on the updated model. Stop when no practically significant Type~I/II factors remain.

\paragraph{Relation to prior approaches.}
Domain randomization~\cite{tobin2017domain}, counterfactual regularization~\cite{mao2021generative}, and active learning~\cite{settles2009active} can be viewed as partial instances of the pipeline: useful correction mechanisms without an explicit factor-level diagnosis stage.
\section{Experiments}
\label{sec:experiments}

We validate SDRS in three settings: controlled diagnosis on dSprites, dense prediction on procedural scenes, and entanglement detection under an imperfect generator.

\subsection{Experiment 1: Controlled Diagnostic Validation}
\label{sec:exp1}

\paragraph{Setup.}
We use dSprites~\cite{dsprites2017}, parameterized by \texttt{shape} (3), \texttt{scale} (6), \texttt{orientation} (40), \texttt{posX} (32), and \texttt{posY} (32). The task is shape classification, so \texttt{shape} is semantic and the remaining factors are nuisance.

\paragraph{Planted biases and splits.}
We construct a biased 30{,}000-image training set with two deficiencies: (i) a shortcut where shape is perfectly correlated with \texttt{posX}; and (ii) a coverage deficiency where only orientations $[0,4]$ are observed. From a balanced pool, we create disjoint \textbf{audit-validation} and \textbf{final test} splits (5{,}000 each). The ANOVA audit and gap assignment are computed on the audit-validation split only; final accuracy is reported on the held-out test split.

\paragraph{Audit results.}
Table~\ref{tab:exp1_anova} identifies both planted issues. \texttt{posX} is significant despite being nuisance ($F{=}2.07$, adjusted $p{<}0.001$). \texttt{orientation} has high sensitivity ($F{=}45.9$) and, crucially, the largest stratified generalization drop on unseen levels, so it is assigned as a Type~I coverage gap under the decision rule in Section~\ref{sec:diagnosis}.

\begin{table}[t]
\centering
\small
\caption{\textbf{Experiment~1: ANOVA audit results (audit-validation split).} Holm-adjusted significance before and after correction.}
\label{tab:exp1_anova}
\vspace{0.3em}
\resizebox{\columnwidth}{!}{%
\begin{tabular}{lrrcrrc}
\toprule
& \multicolumn{2}{c}{\textbf{Before}} & & \multicolumn{2}{c}{\textbf{After}} & \\
\cmidrule{2-3} \cmidrule{5-6}
\textbf{Factor} & $F$ & Sig. & & $F$ & Sig. & \textbf{Gap Type} \\
\midrule
\texttt{shape} & 219.7 & *** & & 8.3 & *** & (semantic) \\
\texttt{scale} & 5.8 & *** & & 26.3 & *** & -- \\
\texttt{orientation} & 45.9 & *** & & 20.3 & *** & Type~I \\
\texttt{posX} & 2.1 & *** & & 1.0 & n.s. & Type~II \\
\texttt{posY} & 1.5 & * & & 1.5 & * & -- \\
\bottomrule
\end{tabular}
}
\end{table}

\paragraph{Correction and baselines.}
SDRS uses 500 counterfactual pairs for \texttt{posX} (Type~II) and 2{,}000 diverse-orientation samples (Type~I), totaling 3{,}000 synthetic images. We fine-tune with Eq.~\eqref{eq:total_loss} and compare against four baselines under the same data budget.

\begin{table}[t]
\centering
\small
\caption{\textbf{Experiment~1: Held-out classification accuracy.} ``Targeted Data'' uses the same diagnosed samples with task loss only. ``Random'' and ``DR'' sample from the full balanced dSprites distribution and are reported as oracle references.}
\label{tab:exp1_acc}
\vspace{0.3em}
\resizebox{\columnwidth}{!}{%
\begin{tabular}{lcc}
\toprule
\textbf{Method} & \textbf{Accuracy} & \textbf{Data Source} \\
\midrule
No synthetic data & 0.499 & -- \\
\midrule
SDRS (task + invariance) & 0.751 & SDRS-diagnosed \\
Targeted Data (task loss only) & 0.790 & SDRS-diagnosed \\
\midrule
Random synthetic$^\dagger$ & 0.824 & Oracle (balanced) \\
Domain randomization$^\dagger$ & 0.851 & Oracle (balanced) \\
\bottomrule
\multicolumn{3}{l}{\footnotesize $^\dagger$ Samples from full balanced data; see text.}
\end{tabular}
}
\end{table}

\paragraph{Analysis.}
Training on diagnosed targeted data improves held-out accuracy from 49.9\% to 79.0\% (+29.1 points). The \texttt{posX} shortcut is removed ($F$: 2.1 $\rightarrow$ 1.0), and unseen-orientation accuracy improves substantially. Adding invariance loss lowers performance relative to task-loss-only training on the same diagnosed data (0.751 vs.\ 0.790), indicating objective competition in the current formulation. We also note \texttt{scale} rises from 5.8 to 26.3, an instance of sensitivity transfer discussed in Section~\ref{sec:disc_transfer}.

\begin{figure*}[t]
\centering
\IfFileExists{sec/figures/sdrs_exp1_results.pdf}{\includegraphics[width=\textwidth]{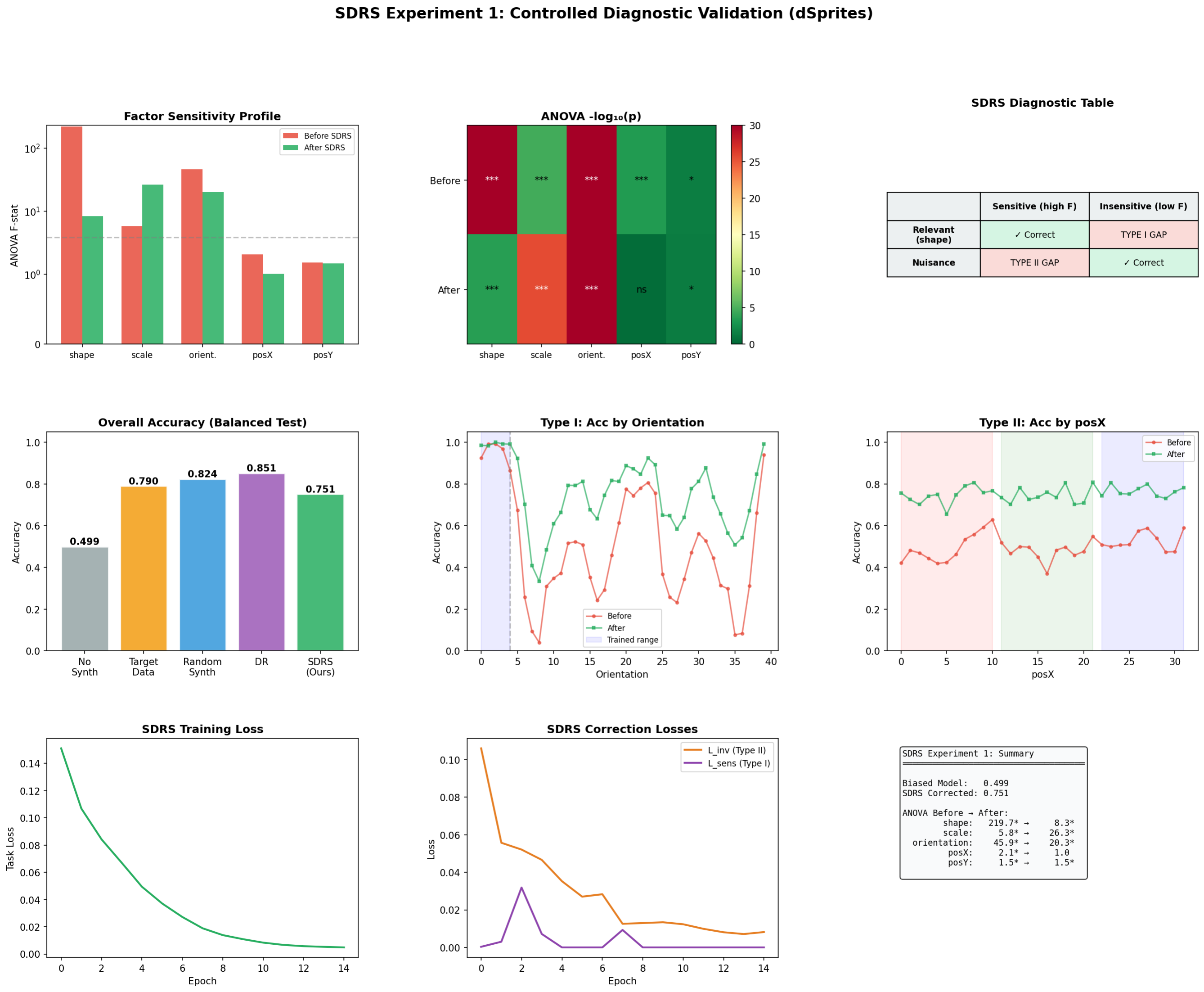}}{\fbox{\parbox[c][0.22\textheight][c]{0.95\textwidth}{\centering Missing figure file}}}
\caption{\textbf{Experiment~1: Controlled diagnostic validation on dSprites.} The audit identifies a \texttt{posX} shortcut (Type~II) and an orientation coverage gap (Type~I). After targeted correction, \texttt{posX} becomes non-significant and orientation sensitivity is reduced by 55.8\%, with improved held-out accuracy. The in-figure diagnostic table illustrates the theoretical taxonomy; operational assignment follows the priority rule in Section~\ref{sec:diagnosis}.}
\label{fig:exp1}
\end{figure*}

\subsection{Experiment 2: Dense Prediction on Procedural Scenes}
\label{sec:exp2}

\paragraph{Setup.}
We build a procedural generator for $128\times128$ RGB scenes with three objects and pixel-perfect segmentation masks (4 classes). Factors are \texttt{light\_dir} (4), \texttt{light\_int} (3), \texttt{bg\_complex} (3), \texttt{obj\_material} (3), \texttt{cam\_angle} (3), and \texttt{occlusion} (3).

\paragraph{Planted biases and splits.}
The biased training set (500 scenes) fixes five factors and varies only \texttt{obj\_material}, inducing nuisance shortcuts and an occlusion coverage gap. We create disjoint balanced \textbf{audit-validation} and \textbf{final test} splits (400 each). ANOVA and diagnosis are computed on the audit-validation split; mIoU is reported on the held-out test split.

\begin{figure*}[t]
\centering
\IfFileExists{sec/figures/sdrs_exp2_results.pdf}{\includegraphics[width=\textwidth]{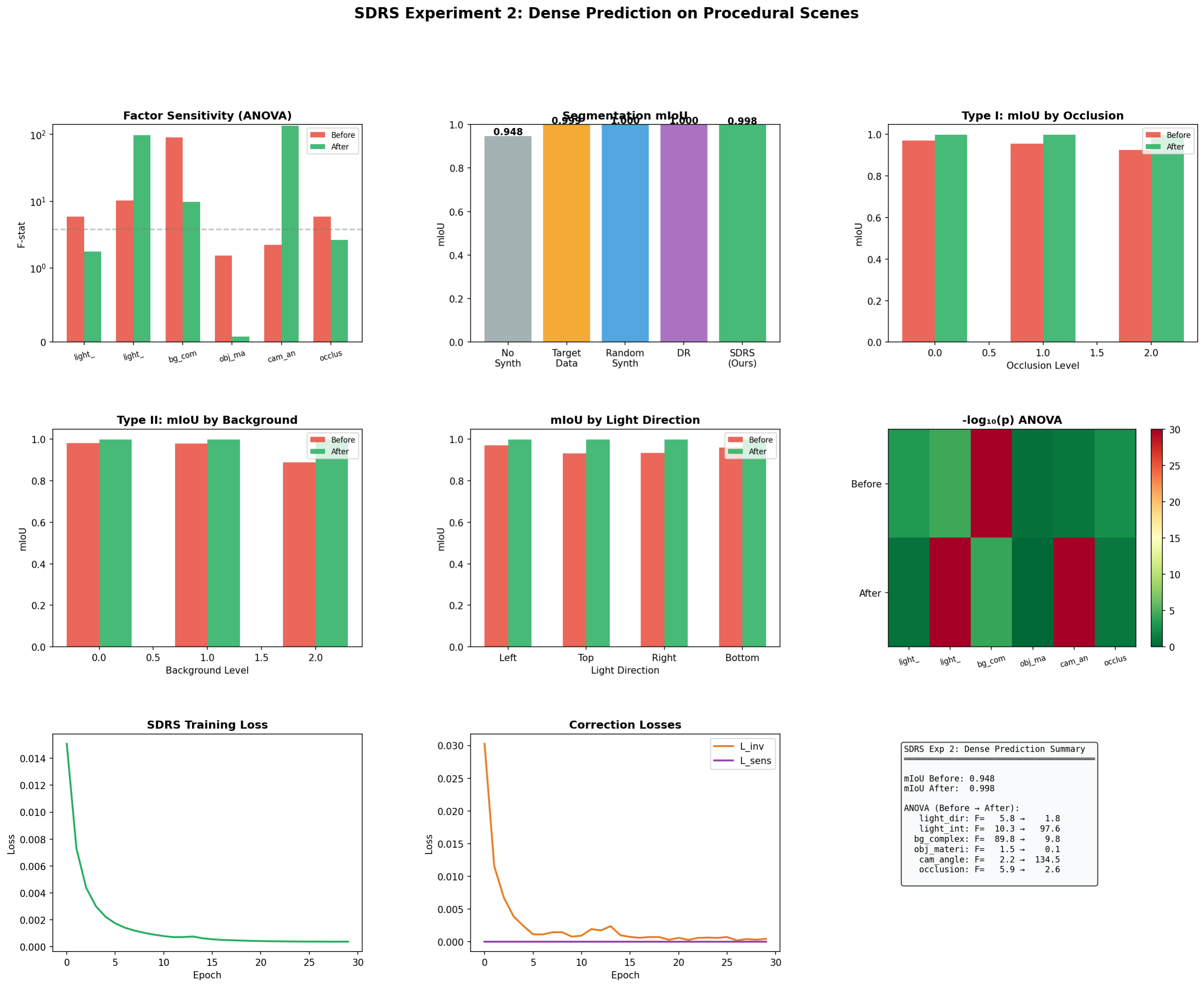}}{\fbox{\parbox[c][0.22\textheight][c]{0.95\textwidth}{\centering Missing figure file}}}
\caption{\textbf{Experiment~2: Dense prediction on procedural scenes.} SDRS reduces the dominant \texttt{bg\_complex} shortcut (89.8 $\rightarrow$ 9.8), closes the occlusion gap, and improves held-out mIoU from 0.948 to 0.998. Task-loss-only training on the same diagnosed data reaches 0.9995, revealing sensitivity transfer under invariance regularization.}
\label{fig:exp2}
\end{figure*}

\paragraph{Audit results.}
Table~\ref{tab:exp2_anova} shows three Type~II gaps and one Type~I gap pre-correction. \texttt{bg\_complex} dominates ($F{=}89.8$), and \texttt{occlusion} is the primary missing-coverage factor.

\begin{table}[t]
\centering
\small
\caption{\textbf{Experiment~2: ANOVA audit and segmentation mIoU.} Audit computed on disjoint audit-validation data; significance is Holm-adjusted.}
\label{tab:exp2_anova}
\vspace{0.3em}
\begin{tabular}{lrrcrrc}
\toprule
& \multicolumn{2}{c}{\textbf{Before}} & & \multicolumn{2}{c}{\textbf{After}} & \\
\cmidrule{2-3} \cmidrule{5-6}
\textbf{Factor} & $F$ & Sig. & & $F$ & Sig. & \textbf{Gap} \\
\midrule
\texttt{light\_dir} & 5.8 & *** & & 1.8 & n.s. & Type~II \\
\texttt{light\_int} & 10.3 & *** & & 97.6 & *** & $\uparrow$ \\
\texttt{bg\_complex} & 89.8 & *** & & 9.8 & *** & Type~II \\
\texttt{obj\_material} & 1.5 & n.s. & & 0.1 & n.s. & -- \\
\texttt{cam\_angle} & 2.2 & n.s. & & 134.5 & *** & $\uparrow$ \\
\texttt{occlusion} & 5.9 & *** & & 2.6 & n.s. & Type~I \\
\bottomrule
\end{tabular}
\end{table}

\paragraph{Correction and baselines.}
SDRS adds 100 targeted correction scenes and 200 counterfactual pairs. All baselines use the same correction budget of 100 additional scenes. All methods fine-tune from the biased model's pretrained weights.

\begin{table}[t]
\centering
\small
\caption{\textbf{Experiment~2: Held-out segmentation mIoU.} SDRS improves mIoU from 0.948 to 0.998. Task-loss-only training on the same diagnosed data achieves 0.9995.}
\label{tab:exp2_miou}
\vspace{0.3em}
\begin{tabular}{lc}
\toprule
\textbf{Method} & \textbf{mIoU} \\
\midrule
No synthetic data & 0.948 \\
\midrule
SDRS (task + invariance) & 0.998 \\
Targeted Data (task loss only) & 0.9995 \\
\midrule
Random synthetic$^\dagger$ & 1.000 \\
Domain randomization$^\dagger$ & 1.000 \\
\bottomrule
\multicolumn{2}{l}{\footnotesize $^\dagger$ Uniformly sampled factor space.}
\end{tabular}
\end{table}

\paragraph{Analysis.}
Diagnosed targeted data closes most of the gap to perfect segmentation (0.948 $\rightarrow$ 0.9995) while resolving the main audited dependencies. With invariance regularization, \texttt{cam\_angle} and \texttt{light\_int} increase sharply (2.2 $\rightarrow$ 134.5 and 10.3 $\rightarrow$ 97.6), indicating sensitivity transfer. Random/DR also reach 1.000 mIoU under the same 100-scene budget because this benchmark saturates quickly under uniform factor coverage. The diagnostic contribution remains: ANOVA identifies why the biased model fails (background shortcut and occlusion coverage failure), information that scalar mIoU alone cannot provide and that transfers to harder benchmarks where ceiling effects are less likely.

\subsection{Experiment 3: Detecting Generator Entanglement}
\label{sec:exp3}

\paragraph{Setup.}
We compare a perfect and an entangled procedural generator for $64\times64$ colored-shape images with factors \texttt{shape}, \texttt{color}, \texttt{size}, \texttt{style}, and \texttt{position}. In the entangled generator, \texttt{style} also changes object size (rough: +30\%, sketchy: $-$21\%).

\begin{figure*}[t]
\centering
\IfFileExists{sec/figures/sdrs_exp3_results.pdf}{\includegraphics[width=\textwidth]{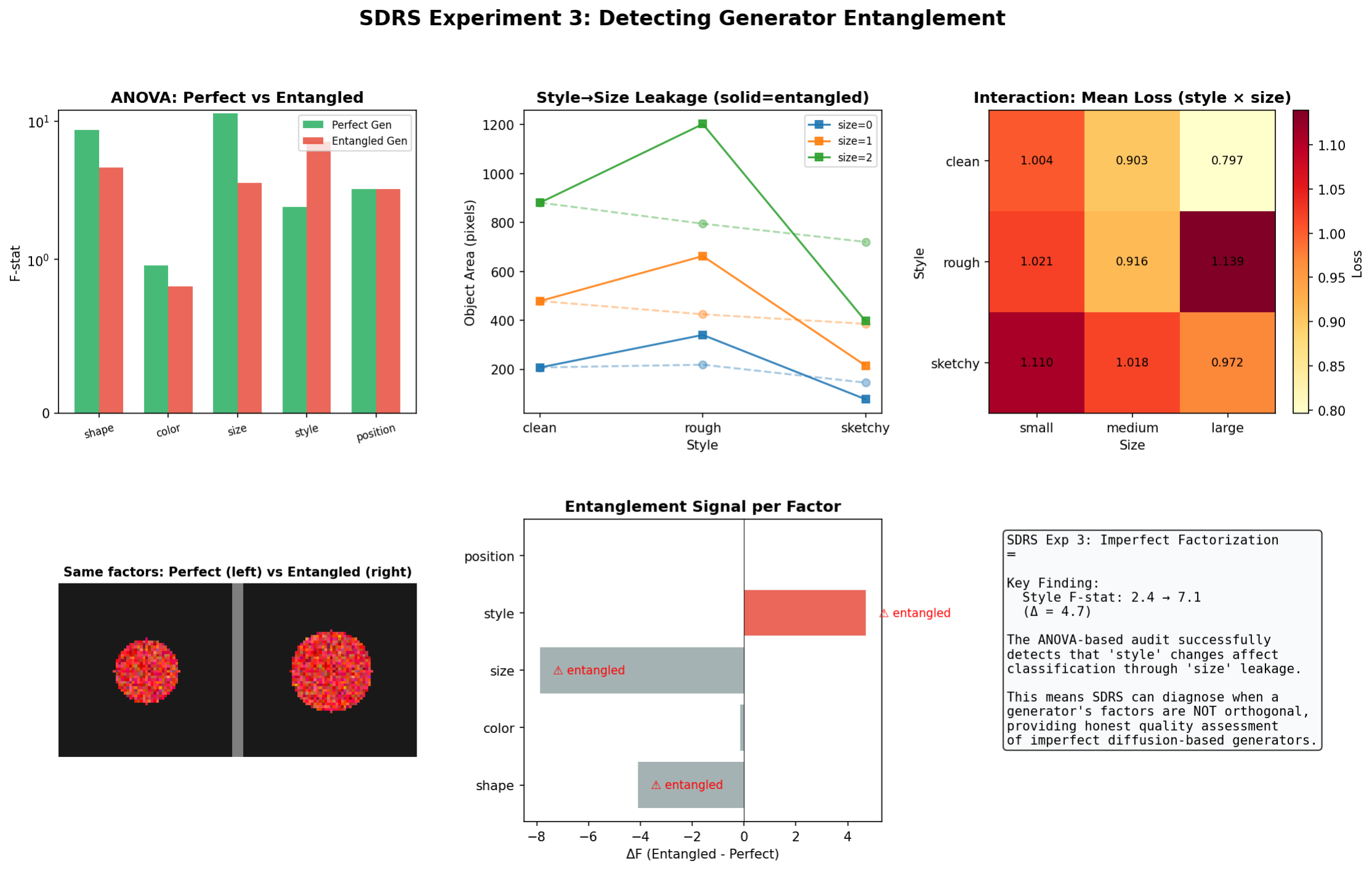}}{\fbox{\parbox[c][0.22\textheight][c]{0.95\textwidth}{\centering Missing figure file}}}
\caption{\textbf{Experiment~3: Detecting generator entanglement.} Comparing perfect and entangled generators, the audit detects style$\rightarrow$size leakage: \texttt{style} rises from 2.4 to 7.1 while \texttt{size} drops from 11.4 to 3.6.}
\label{fig:exp3}
\end{figure*}

\paragraph{Results and implications.}
Table~\ref{tab:exp3} shows the expected shift under entanglement: \texttt{style} increases by +4.7 and \texttt{size} decreases by $-$7.9. This indicates that the ANOVA audit can also evaluate generator quality by flagging cross-factor contamination from empirical sensitivity profiles.

\begin{table}[t]
\centering
\small
\caption{\textbf{Experiment~3: Entanglement detection.} ANOVA $F$-statistics for perfect vs. entangled generators (audit-validation split).}
\label{tab:exp3}
\vspace{0.3em}
\begin{tabular}{lrrc}
\toprule
\textbf{Factor} & $F$ \textbf{(Perfect)} & $F$ \textbf{(Entangled)} & $\Delta F$ \\
\midrule
\texttt{shape} & 8.7 & 4.6 & $-4.1$ \\
\texttt{color} & 1.0 & 0.8 & $-0.1$ \\
\texttt{size} & 11.4 & 3.6 & $-7.9$ \\
\texttt{style} & 2.4 & 7.1 & $+4.7$ \\
\texttt{position} & 3.2 & 3.2 & $\phantom{+}0.0$ \\
\bottomrule
\end{tabular}
\end{table}
\section{Discussion}
\label{sec:discussion}

\subsection{The Diagnostic as the Primary Contribution}
\label{sec:disc_diagnostic}

Across all three experiments, the ANOVA-based audit identifies the dominant vulnerabilities on disjoint audit-validation splits. In Experiment~1, it finds a \texttt{posX} shortcut and an orientation coverage failure; in Experiment~2, it identifies \texttt{bg\_complex} as the dominant shortcut and \texttt{occlusion} as the primary missing-coverage factor; in Experiment~3, it detects style$\rightarrow$size entanglement ($\Delta F = +4.7$) without ground-truth entanglement labels.

The practical value is decomposition: SDRS reports \emph{which factors} drive failure and \emph{which correction type} they require, rather than only aggregate scores such as accuracy or mIoU.

\subsection{Sensitivity Transfer: An Identified Open Problem}
\label{sec:disc_transfer}

In Experiment~2, the invariance penalty reduces dependence on targeted nuisance factors (\texttt{bg\_complex}: $89.8 \rightarrow 9.8$; \texttt{light\_dir}: $5.8 \rightarrow 1.8$) but increases sensitivity to non-targeted factors (\texttt{cam\_angle}: $2.2 \rightarrow 134.5$; \texttt{light\_int}: $10.3 \rightarrow 97.6$). We term this \emph{sensitivity transfer}. A plausible mechanism is representational reallocation: suppressing some nuisance directions in $\Phi_w^{(l)}$ leaves capacity that is repurposed toward other nuisance cues.

This explains why the current correction objective does not consistently outperform task-loss-only training on SDRS-diagnosed data (Table~\ref{tab:exp1_acc}: 0.790 vs.\ 0.751; Table~\ref{tab:exp2_miou}: 0.9995 vs.\ 0.998). In this version, the strongest contribution is the diagnostic stage; correction-loss design remains open.

\subsection{Limitations}
\label{sec:limitations}

\paragraph{Factor coverage and interaction assumptions.}
SDRS audits only factors explicitly exposed by the generator. Unmodeled variation remains outside scope. The current one-way ANOVA focuses on marginal effects; interaction-heavy settings may require multi-way ANOVA/ANCOVA and larger designs.

\paragraph{Generator realism and factorization quality.}
Experiments use procedural generators with perfect (Experiments~1--2) or controlled (Experiment~3) orthogonality. Diffusion generators (e.g., ControlNet~\cite{zhang2023adding}) provide richer realism but weaker factor separation; broader validation on diffusion-generated data is still needed.

\paragraph{Correction loss robustness.}
As discussed in Section~\ref{sec:disc_transfer}, per-factor invariance penalties can transfer sensitivity across nuisance factors. Improving correction losses beyond task-loss-only training on diagnosed data is a central next step.
\section{Conclusion}
\label{sec:conclusion}

We have presented SDRS, a framework that connects the statistical theory of Design of Experiments to synthetic data generation for computer vision. The core idea is to treat the downstream model as a system under investigation and the synthetic generator as an experimental apparatus, using structured factorial designs to efficiently probe the model's factor-sensitivity profile.

The ANOVA-based audit at the heart of SDRS provides a decomposed, per-factor diagnostic that identifies both spurious shortcuts (Type~II gaps) and missing capabilities (Type~I gaps). Across three experiments, the audit correctly identifies planted biases, tracks their evolution through correction, and detects cross-factor entanglement in imperfect generators. Data targeted by the diagnostic produces substantial performance gains over no-synthetic baselines (+29.1 pp in classification, +5.0 points mIoU in segmentation).

Our investigation also identified an open problem: per-factor invariance penalties can transfer sensitivity between nuisance factors rather than eliminating it, a phenomenon we term \emph{sensitivity transfer}. This finding suggests that the correction phase of synthetic data pipelines requires more careful design than the field has so far recognized, and that the diagnostic framework we propose is a necessary first step---one must understand what is wrong before one can fix it.

We believe the central insight of this work---that synthetic data pipelines should be designed as structured experiments, not random sampling processes---provides a productive direction for the community, and we hope the SDRS diagnostic will serve as a practical tool for both evaluating models and assessing the quality of synthetic generators.

{
    \small
    \bibliographystyle{ieeenat_fullname}
    \bibliography{main}
}

\clearpage
\hypersetup{pageanchor=false}
\setcounter{page}{1}
\maketitlesupplementary

\appendix
\renewcommand{\thetable}{S\arabic{table}}
\renewcommand{\thefigure}{S\arabic{figure}}
\renewcommand{\theequation}{S\arabic{equation}}

\section{Model Architectures}
\label{sec:supp_arch}

\paragraph{Experiment 1: dSprites CNN.}
The downstream model is a four-layer CNN:
\texttt{Conv2d(1,32,3,stride=2,pad=1)} $\rightarrow$ BN $\rightarrow$ ReLU
$\rightarrow$ \texttt{Conv2d(32,64,3,2,1)} $\rightarrow$ BN $\rightarrow$ ReLU
$\rightarrow$ \texttt{Conv2d(64,128,3,2,1)} $\rightarrow$ BN $\rightarrow$ ReLU
$\rightarrow$ \texttt{Conv2d(128,256,3,2,1)} $\rightarrow$ BN $\rightarrow$ ReLU
$\rightarrow$ \texttt{AdaptiveAvgPool2d(1)} $\rightarrow$ \texttt{Linear(256,128)}
$\rightarrow$ ReLU $\rightarrow$ \texttt{Linear(128,3)}.
Feature extraction for the invariance loss uses the 128-dimensional output
of the penultimate linear layer. Total parameters: $\sim$430K.

\paragraph{Experiment 2: Small U-Net.}
We use a four-stage encoder-decoder with skip connections. Each encoder stage
uses \texttt{Conv2d $\rightarrow$ BN $\rightarrow$ ReLU $\rightarrow$ Conv2d $\rightarrow$ BN $\rightarrow$ ReLU},
with channel progression
$3\rightarrow32\rightarrow64\rightarrow128\rightarrow256$ in the encoder and
$256\rightarrow128\rightarrow64\rightarrow32$ in the decoder. Upsampling uses
bilinear interpolation followed by a convolutional block. The segmentation
head is a $1\times1$ convolution mapping 32 channels to 4 classes. Feature
extraction for the invariance loss uses the third encoder stage output
($32\times32\times128$). Total parameters: $\sim$1.9M.

\paragraph{Experiment 3: Tiny CNN.}
\texttt{Conv2d(3,16,3,2,1)} $\rightarrow$ ReLU
$\rightarrow$ \texttt{Conv2d(16,32,3,2,1)} $\rightarrow$ ReLU
$\rightarrow$ \texttt{Conv2d(32,64,3,2,1)} $\rightarrow$ ReLU
$\rightarrow$ \texttt{AdaptiveAvgPool2d(1)} $\rightarrow$ \texttt{Linear(64,3)}.
Total parameters: $\sim$25K.

\section{Training Hyperparameters}
\label{sec:supp_hyper}

All experiments use Adam with default momentum
($\beta_1{=}0.9$, $\beta_2{=}0.999$) and cosine-annealing learning-rate
schedules. \Cref{tab:supp_hyper} summarizes the full settings.

\begin{table}[t]
\centering
\small
\caption{Training hyperparameters across experiments.}
\label{tab:supp_hyper}
\begin{tabular}{lccc}
\toprule
& \textbf{Exp 1} & \textbf{Exp 2} & \textbf{Exp 3} \\
\midrule
Learning rate & $3 \times 10^{-4}$ & $1 \times 10^{-3}$ & $1 \times 10^{-3}$ \\
Batch size & 256 & 32 & 64 \\
Epochs (biased) & 15 & 30 & 15 \\
Epochs (correction) & 15 & 30 & -- \\
$\lambda$ (Eq.~\eqref{eq:total_loss}) & 0.5 & 0.5 & -- \\
Sensitivity margin & 3.0 & 3.0 & -- \\
Inv.\ pairs per batch & 32 & 32 & -- \\
Sens.\ pairs per batch & 64 & -- & -- \\
\bottomrule
\end{tabular}
\end{table}

\section{Fractional Factorial Design Details}
\label{sec:supp_doe}

For Experiment~1 ($k{=}5$ factors), we use a $2^{5-2}_{\mathrm{IV}}$
fractional factorial with generators $D{=}AB$ and $E{=}AC$, yielding 8 runs.
Each factor is mapped to two levels: low = first quartile of the factor range
and high = last quartile. For example, \texttt{orientation} has 40 levels;
low = $\{0,\ldots,9\}$ and high = $\{30,\ldots,39\}$.

In practice, we run the ANOVA audit on the balanced audit-validation split
(5,000 images for Experiment~1 and 400 images for Experiment~2), rather than
on a dedicated 8-point probe set. This provides higher statistical power while
preserving the designed-intervention logic. Section~3.1 of the main paper
therefore describes the \emph{minimal} protocol when a balanced evaluation set
is unavailable, whereas the experiments report results under the more favorable
condition of balanced evaluation data.

\section{Planted Bias Details}
\label{sec:supp_bias}

\paragraph{Experiment 1.}
The biased training set enforces:
\begin{itemize}[leftmargin=*,itemsep=1pt]
\item \texttt{shape}=0 (square) $\Rightarrow$ \texttt{posX} $\in [0, 10]$
\item \texttt{shape}=1 (ellipse) $\Rightarrow$ \texttt{posX} $\in [11, 21]$
\item \texttt{shape}=2 (heart) $\Rightarrow$ \texttt{posX} $\in [22, 31]$
\item \texttt{orientation} $\in [0, 4]$ (out of 40 levels)
\end{itemize}
Verification from training statistics: shape=0 has mean \texttt{posX}=5.0,
shape=1 has mean \texttt{posX}=16.0, and shape=2 has mean \texttt{posX}=26.5.
The biased model reaches 100\% accuracy on the biased training set,
confirming that the shortcut is trivially learnable.

\paragraph{Experiment 2.}
The biased training set fixes: \texttt{light\_dir}=0 (frontal),
\texttt{light\_int}=1 (normal), \texttt{bg\_complex}=0 (plain),
\texttt{cam\_angle}=0 (eye-level), and \texttt{occlusion}=0 (none).
Only \texttt{obj\_material} varies across its 3 levels.

\section{Holm--Bonferroni Correction Procedure}
\label{sec:supp_holm}

Given $k$ factor-wise ANOVA tests with ordered raw p-values
$p_{(1)} \leq p_{(2)} \leq \cdots \leq p_{(k)}$, the Holm--Bonferroni rule
rejects $H_{0,(i)}$ if
\begin{equation}
p_{(i)} \leq \frac{\alpha}{k-i+1},
\end{equation}
proceeding sequentially and stopping at the first non-rejection.
For Experiment~1 ($k{=}5$, $\alpha{=}0.05$), thresholds are
$0.01$, $0.0125$, $0.0167$, $0.025$, and $0.05$.

In Table~1 of the main paper, all factors except \texttt{posY} have
$p<0.001$, so they remain significant after adjustment. \texttt{posY} has
$p\approx0.033$ and, as the largest p-value, is tested against $0.05$;
it remains significant.

\section{Sensitivity Transfer: Additional Analysis}
\label{sec:supp_transfer}

\Cref{tab:supp_transfer} reports complete before/after ANOVA profiles,
including factors that exhibit sensitivity transfer (marked with $\uparrow$).

\begin{table}[t]
\centering
\small
\caption{Complete ANOVA $F$-statistics before and after SDRS correction.
Factors marked $\uparrow$ exhibit sensitivity transfer.}
\label{tab:supp_transfer}
\resizebox{\columnwidth}{!}{%
\begin{tabular}{lrrrl}
\toprule
\multicolumn{5}{c}{\textbf{Experiment 1}} \\
\midrule
Factor & $F_{\text{before}}$ & $F_{\text{after}}$ & $\Delta\%$ & Note \\
\midrule
\texttt{shape} & 219.7 & 8.3 & $-96.2$ & semantic \\
\texttt{scale} & 5.8 & 26.3 & $+355.9$ & $\uparrow$ transfer \\
\texttt{orient.} & 45.9 & 20.3 & $-55.8$ & Type I corrected \\
\texttt{posX} & 2.1 & 1.0 & $-50.8$ & Type II corrected \\
\texttt{posY} & 1.5 & 1.5 & $-2.4$ & unchanged \\
\midrule
\multicolumn{5}{c}{\textbf{Experiment 2}} \\
\midrule
Factor & $F_{\text{before}}$ & $F_{\text{after}}$ & $\Delta\%$ & Note \\
\midrule
\texttt{light\_dir} & 5.8 & 1.8 & $-69.9$ & Type II corrected \\
\texttt{light\_int} & 10.3 & 97.6 & $+849.3$ & $\uparrow$ transfer \\
\texttt{bg\_complex} & 89.8 & 9.8 & $-89.1$ & Type II corrected \\
\texttt{obj\_mat.} & 1.5 & 0.1 & $-95.2$ & unchanged \\
\texttt{cam\_angle} & 2.2 & 134.5 & $+5977.8$ & $\uparrow$ transfer \\
\texttt{occlusion} & 5.9 & 2.6 & $-55.1$ & Type I corrected \\
\bottomrule
\end{tabular}
}
\end{table}

The transfer pattern is consistent across both experiments: factors not
explicitly targeted by the invariance penalty can increase in sensitivity
after correction. In Experiment~1, \texttt{scale} increases from 5.8 to 26.3.
In Experiment~2, both \texttt{cam\_angle} and \texttt{light\_int} increase
substantially. In contrast, targeted factors decrease as intended.

This behavior is consistent with a fixed-capacity representation hypothesis:
suppressing targeted nuisance directions can free representational capacity
that is then reallocated to other available cues, including non-targeted
nuisance factors.

\section{Entanglement Generator Specification}
\label{sec:supp_entangle}

The entangled generator modifies rendered object size as a function of
\texttt{style}:
\begin{itemize}[leftmargin=*,itemsep=1pt]
\item \texttt{style}=0 (clean): $\text{size}_{\text{rendered}} = \text{size}_{\text{base}}$
\item \texttt{style}=1 (rough):
$\text{size}_{\text{rendered}} = \text{size}_{\text{base}} \times (1 + \epsilon)$, with $\epsilon = 0.3$
\item \texttt{style}=2 (sketchy):
$\text{size}_{\text{rendered}} = \text{size}_{\text{base}} \times (1 - 0.7\epsilon)$, with $\epsilon = 0.3$
\end{itemize}
The perfect generator sets $\epsilon=0$ for all styles.
Base sizes are: small = 8 px, medium = 12 px, and large = 16 px
(radius/half-side) on a $64\times64$ canvas.

\end{document}